# S$^2$DEVFMAP: Self-Supervised Learning Framework with Dual Ensemble Voting Fusion for Maximizing Anomaly Prediction in Timeseries

SARALA.M.NAIDU*

Dept. of Innovation, Design and Technology, Mälardalen University, Västerås, Sweden

Power Grids, Grid Integeration, Hitachi Energy, Västerås, Sweden

Ning Xiong

Dept. of Innovation, Design and Technology, Mälardalen University, Västerås, Sweden

Anomaly detection plays a crucial role in industrial settings, particularly in maintaining the reliability and optimal performance of cooling systems. Traditional anomaly detection methods often face challenges in handling diverse data characteristics and variations in noise levels, resulting in limited effectiveness. And yet traditional anomaly detection often relies on application of single models. This work proposes a novel, robust approach using five heterogeneous independent models combined with a dual ensemble fusion of voting techniques. Diverse models capture various system behaviors, while the fusion strategy maximizes detection effectiveness and minimizes false alarms. Each base autoencoder model learns a unique representation of the data, leveraging their complementary strengths to improve anomaly detection performance. To increase the effectiveness and reliability of final anomaly prediction, dual ensemble technique is applied. This approach outperforms in maximizing the coverage of identifying anomalies. Experimental results on a real-world dataset of industrial cooling system data demonstrate the effectiveness of the proposed approach. This approach can be extended to other industrial applications where anomaly detection is critical for ensuring system reliability and preventing potential malfunctions.

## 1 INTRODUCTION

Ensuring the reliability of cooling systems is crucial in power plants. Traditional, single-model anomaly detection application can miss diverse issues. This work proposes a novel approach using heterogeneous models, each capturing different aspects of behavior. Further, a dual ensemble voting fusion strategy combines the diverse outputs, maximizing detection effectiveness and minimizing false alarms.

### 1.1 Background

Industrial cooling systems are crucial for power transmission, but prone to failures as traditional maintenance relies on preplanned schedules that are based on the FMECA (Failure Mode Effects and Criticality Analysis) relevant failure modes, as an input for the analysis. Thus, the proposed approach explores anomaly detection for predictive maintenance, aiming to reduce downtime, enhance reliability, and minimize costs. This work uses diverse models and a fusion technique to detect expand the anomaly coverage across various components for earlier issue identification and more efficient maintenance. This approach complements existing practices in SCADA with alarm systems, offering a proactive and data-driven solution for more reliable cooling systems. The following are the challenges of the existing practices:

I. Complexity and Maintenance:
   a. Knowledge engineering bottleneck: Defining and maintaining many rules, can be time-consuming and require expert knowledge.
   b. Debugging and testing: Complex rule interactions can be hard to debug and test.
II. Restricted Intelligence and Scalability:
   c. Limited to predefined rules: The system can only make decisions based on its explicitly programmed rules, hindering its ability to handle ambiguity or novel situations.
   d. Scalability issues: Adding new rules can become cumbersome in large-scale applications.

* Corresponding Author. The work has been supported by the Swedish Foundation for Strategic Research (Project: ID20-0019)

Thus, data-driven approach with deep learning on the operation data to detect and identify anomalies upfront using the streaming data would help the easy adaptation for predictive maintenances approach, by addressing:

I. Reduce false positives/negatives: Deep learning (DL) can be trained to distinguish subtle nuances and patterns, potentially leading to fewer incorrect classifications, or missed opportunities.
II. Predictive insights by analyzing data to predict future outcomes, enabling proactive actions and improved decision-making.
III. Data availability and quality: DL models rely heavily on large amounts of high-quality data that is aggregated from existing SCADA system.
IV. Dynamic adaptation: DL models can continuously learn and adapt to new data, automatically adjusting rules as situations change as in conditions that constantly evolve.

Thus, the key contributions are as follows:

**C1:** A Novel framework for Proactive Anomaly Detection for Predictive Maintenance in industrial cooling systems for power transmission. This provokes the shifts from traditional scheduled maintenance to predictive maintenance and has the potential to significantly reduce downtime, enhance reliability, and minimize costs.

**C2:** Heterogeneous Models: to leverage diverse models to capture various anomaly patterns across the cooling system and improves overall detection capabilities compared to single-model approaches.

**C3:** Dual Ensemble Fusion of Voting Techniques: the approach aims to maximize detection accuracy and minimize false alarms, by combining the outputs from two ensembles voting methods.

**C4:** Focus on real data from Industrial Cooling Systems: While anomaly detection for industrial settings is not entirely new, this work specifically focuses on data from a real industrial cooling system.

## 1.2 Related Work

Since there is no exact or similar approach in any literature paper to compare either results or the methods, the in general anomaly detection using Ensemble methods and its application in Power Industry is discussed.

### 1.2.1 Anomaly detection in time series

Rare but critical anomalies present challenges for training models, as the data grows. Unsupervised methods, like autoencoders, gain popularity for their anomaly detection capability. While LSTMs offer better results [1], their complexity is expensive. Various other DL models, including LSTMs, CNNs, and attention layers, showcase success in identifying power plant issues [2-4].

### 1.2.2 Ensemble Methods

Ensemble learning (EL) combines AI algorithms for accuracy and robustness, addressing individual model limitations across datasets. However, training and testing multiple models can be computationally expensive. Mathematically, ensembles balance bias and variance, improving consistency and accuracy, by combining diverse strengths, leading to better generalization. EL applications and state-of-the-art techniques are reviewed in [5], [6] and show the use of any one of the voting mechanisms applied across different use-cases.

### 1.2.3 Ensembled Fusing of Voting methods

An EL systems uses an aggregate function 'f' to combine the base models 'M' to predict the output of an unsupervised dataset 'D' with size n and feature dimension m.

$$D = (x_i), 1 \leq i \leq n, x_i \in R^m \quad (1)$$
$$y_i = f(M_1, M_2....M_k) \quad (2)$$

where M is the heterogeneous model from 1 to k (here 5). Despite challenges in creating model diversity, ensembles excel by merging outputs (voting/meta-learning) and leveraging their strengths. The [5] showcase its growing preference across domains, highlighting its ability to create more robust and superior models than single learners. While common methods like voting/stacking are used, simply averaging deep learning models can be risky due to biases. Our novel "dual ensemble voting" tackles this, aiming for stronger anomaly detection.

### 1.2.4 Application in Power Industry application

Energy and power domains leverage ensemble learning (EL) for various tasks like anomaly detection and forecasting. References [7-9] showcase applications in smart grid intrusion detection, building energy consumption, gas turbine anomaly detection, and electricity consumption forecasting using diverse techniques like bagging, boosting, stacking, and deep AEs. These methods address limitations of unsupervised learning by combining diverse base learners, leading to more robust and accurate results. This paper proposes a unique framework leveraging heterogeneous base models and a strategic combination strategy for reliable anomaly detection, overcoming challenges associated with unsupervised learning.

## 2 OVERALL METHODOLOGY AND FRAMEWORK

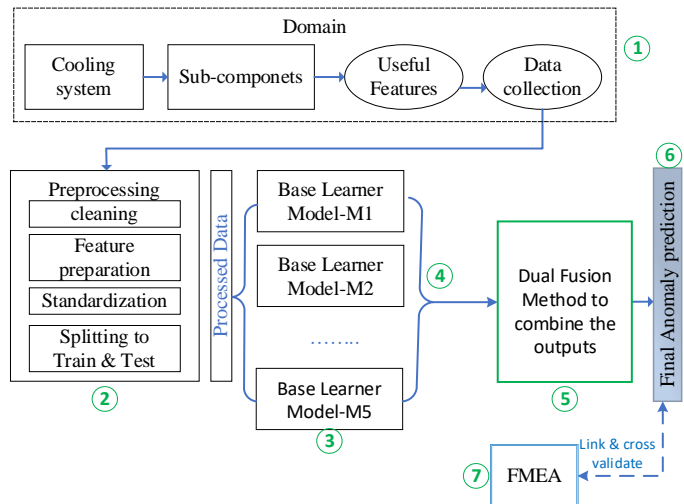

Figure 1: [a] The Proposed Framework.

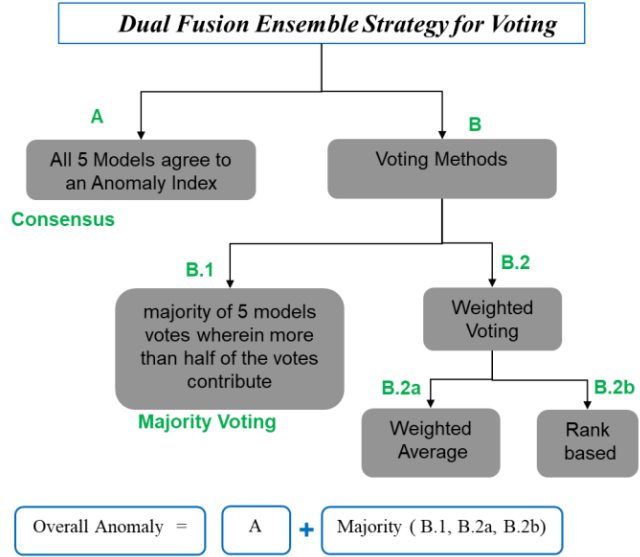

Figure 1: [b] Dual Fusion strategy

The overall framework shown in Figure 1 [a] has seven main steps: (i) Domain Data aggregation, (ii) data Preprocessing, (iii) independent Heterogenous models training and testing, (iv) individual model testing for mae, (v) dual fusion strategy as in Figure 1 [b], (vi) finalization of the overall quantity of risks], and (vii) Linking with FMEA that was prepared at design time.

## 2.1 Data Aggregation and preprocessing

Sensors collect data at varying intervals (2020-2023) including temperature, conductivity, and more. Preprocessing cleans, resamples, and scales the data (accounting for diverse ranges). To prevent leakage, 2020 data is used for anomaly validation, while 2021-2023 is split 80/20 for training/testing. Finally, Equation (3) normalizes the data for equal feature weights.

$$z_i = \frac{x_i - \mu}{\sigma} \quad (3)$$

Where mean $\mu = \frac{1}{N}\sum_{i=1}^{N}(x_i)$ and the standard deviation σ is given by $\sigma = \sqrt{\frac{1}{N}\sum_{i=1}^{N}(x_i - \mu)^2}$.

## 2.2 Base Models

Diverse base models, each with optimized hyperparameters from Table 1, undergo cross-fold validation. Their performance is measured and recorded as individual MAE scores. The details on the individual models and their architecture are presented in our extended work in [2] where in the Meta learning-based ensemble is covered. Compare to the approach presented in this work, in [2], there is two level of training involved and thus has a longer effort. When it comes to practicality for model deployments in real system, the proposed dual ensemble of voting fusion can be applied without compromising the detection of risks.

Table 1: Hyperparameters used in cross fold validation on Base Models

| Hyperparameters | Range of value configured |
| --- | --- |
| TimeSteps | 1 hr to 24 hr |
| Batch size | 16 to 64 |
| Learning rate | 0.01 to 0.0001 |
| Epochs | 50 to 500 with callback to monitoring the loss, with patience=5 |
| Optimizer | Adam |

## 2.3 Dual Voting Fusions Strategy

Detecting an anomaly or normal instance is treated as a two-class problem, with classes as 0 for Normal, 1 for Anomaly. For every instance of inputs x to be predicted, the model M predicts the class ŷ for each input x. The predictors 'x' may be the offline batch data, or online data stream. Final prediction of the total anomaly instances is determined from the individual models by dual fusion of voting approaches, described in the following section.

### 2.3.1 All models agree / consensus-based method:

Referred here as consensus-based voting involves quantifying all anomaly points that are commonly identified by all models. Here, the label ŷ is predicted when each learner base model $M_j$, (where j is number of models, here 1 to 5), predicts the same label, as given by Eq. (4) and the number of anomalies from this approach $N_a$ is the sum of the anomalies in the sequence as given by Eq. (5).

$$ŷ = \text{common}\{M_1(x), M_2(x),..., M_n(x)\} \quad (4)$$
$$N_a = \sum \text{common}\{M_1(x), M_2(x),..., M_n(x)\} \quad (5)$$

### 2.3.2 Other Voting based methods:

For higher reliable prediction, here the results of other voting techniques are fused to combine the prediction from the group of base learners. There are three variations in the approach of applying the voting. Let, $W_j$ be the weight assigned to the models (1 to 5) based on its performance metric of the individual model's mae score.

B.1 <u>Hard or Majority voting</u>: It is the simple form of voting where the label ŷ is predicted via majority voting of each learner model $M_j$, as given by Eq. (6), wherein more than half of the models votes contribute.

$$ŷ = \text{majority}\{M_1(x), M_2(x),.... M_j(x)\} \quad (6)$$

wherein more than half the number of models contribute. The number of anomalies is $N_{b.1}$ is as per Eq. (9).

B.2 <u>Weighted Average Voting</u>: Here, the calculated weight $W_j$ is associated to the model $M_j$, as per n Eq. (7). The final predictions are based on Eq. (8) and total number of anomalies $N_{b.2a}$ is as per Eq. (9).

$$W_{jε(1..5)} = 1 - \text{mae}(M_j) \quad (7)$$

$$ŷ = \begin{cases} 1, & \frac{1}{\sum W}\sum_{j=1}^{5} W_j M_j(x) > 0.5 \\ 0, & otherwise \end{cases} \quad (8)$$

$$N_b = \sum ŷ \quad (9)$$

Here, the weight of each model is set proportional to the mae score, the lower the score, gets a higher weight. Model with a higher performance have a large share of the vote.

B.3 <u>Rank Voting</u>: Instead of weights for each model, another method that can be applied is the rank-based voting. Here the worst model (based on the model's performance mae score, the higher the score the worst is the model) gets the rank 1, and the best model gets the rank 5. The weighted rank $WR_j$ is derived by summing up the ranks and dividing each of the rank by the total value, as shown in Eq. (10).

$$RW_{jε(1..5)} = R_j / \sum_{j=1}^{5} R_j \quad (10)$$

Equation (11) represents the final prediction which is the sum of the product of each weight and the individual model output prediction. The Number of anomalies $N_{b.2b}$ is derived from Eq. (9).

$$ŷ = \begin{cases} 1, & \sum_{j=1}^{5} RW_j M_j(x) > 0.5 \\ 0, & otherwise \end{cases} \quad (11)$$

*As the industrial data is noisy, contain inherent biases and anomalies may manifest in various forms., single* models can be overly sensitive to these issues, leading to false positives or negatives. By combining predictions from multiple models with different learning styles with this choice of dual fusion strategy, can be more robust to noise and data biases, resulting in more reliable anomaly detection. It also allows for integrating multiple data sources, providing a more complete picture of the industrial cooling system's performance.

## 2.4 Ensemble for Final Prediction

Since, the goal is to find the anomaly instances and quantify the number of anomalies, the maximum probable anomalies is determined by the fusion of the voting methods described. Thus, the total number of anomaly instances detected 'N' will be as given in Eq. (12).

$$N = N_a + N_b \quad (12)$$

Where $N_a$ and $N_b$ are from (5) and (9) respectively. And $N_b$ as shown in Figure 1[b] is the majority of $N_{b.1}$, $N_{b.2a}$, and $N_{b.2a}$, that is a fusion of different voting methods described.

With respect to the ensemble review in in [5], [6] and others, most of the different applications of ensemble is based either only weighted or voting mechanism only and moreover the choice on the number of models is limited to three models. Even though it takes a longer time and computational power, as the number of models increases, in this application, to evaluate the performance 5 models are selected. The contribution C1 is mainly because the primary objective of the proposed approach is to maximize anomaly identification coverage effectively. While existing methods may employ limited models for anomaly detection, the use of five diverse independent models coupled with a dual ensemble fusion suggests a more comprehensive and robust strategy for detecting anomalies across various scenarios. The effectiveness of the approach is demonstrated through experimentation on real-world datasets, further supporting its novelty. And by addressing challenges faced by traditional anomaly detection methods, particularly in handling diverse data characteristics, the proposed approach fills a gap in the existing literature. Secondly, the emphasis on proactive anomaly detection for predictive maintenance in industrial cooling systems is itself a novel aspect. Shifting from traditional scheduled maintenance to predictive maintenance can significantly enhance reliability, reduce downtime, and minimize costs. While detailed comparisons with existing methods could further strengthen the claim of novelty, the unique combination of models, context of application in industrial cooling systems, superior performance, and potential paradigm shift towards predictive maintenance contribute to establishing the novelty of the proposed methodology. For all experiments the same fixed random seed was chosen to ensure reproducibility.

## 3 EXPERIMENT RESULTS AND DISCUSSIONS

### 3.1 Base-learner model results

Each of the base model were trained independently with own set of hyperparameters to achieve the best results with the validation loss being monitored. For validation, a random month's data of the year-2020 was taken, and here it was September-2020 data used in the representation of the discussion of the results. Table 2 shows the mae and corresponding loss plots on new unseen validation data in Figure 2 [a to e].

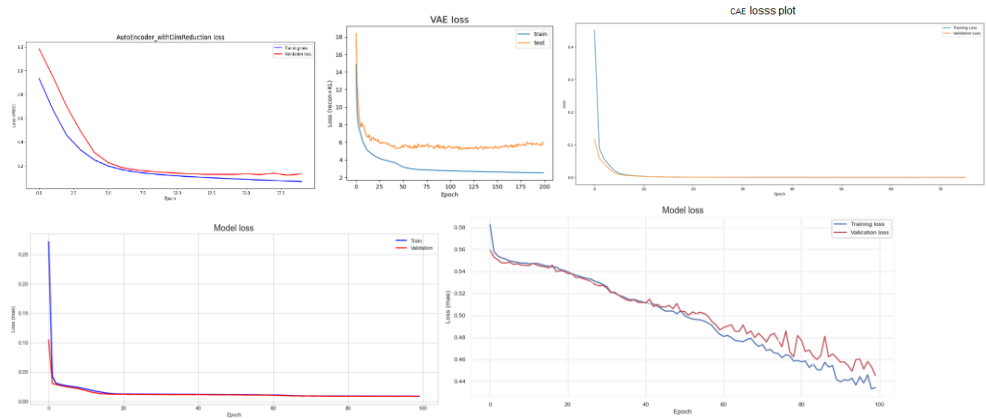

Figure 2: [a] AE Loss Plot ; [b] VAE Loss Plot ; [c] CAE Loss Plot ; [d] LSTM-AE Loss Plot ; [e] LSTM-VAE Loss

Table 2: Results of individual Base Models

| Model Name | Test Data MAE |
|---|---|
| AE | 0.43 |
| VAE | 0.453 |
| CAE | 0.007 |
| LSTM-AE | 0.116 |
| LSTM-VAE | 0.484 |

Issues like overfitting or underfitting in the model can be spotted by analyzing training & validation loss plot. Looking at Table 2, the CAE model emerges as the champion, with lowest MAE.

### 3.2 Ensemble of Fusion with dual voting results

Each of the base learner autoencoder variants exhibited a range of reconstruction quality in pre-training achieving MAE metrics ranging from 0.484 to 0.007. A lower MAE indicates that the autoencoder is performing well at reconstructing the original input data, while the higher MAE signifies that it is struggling to accurately reconstruct the input data, despite optimal training process. This could be because the specific data this autoencoder encountered during training might have been inherently more challenging to reconstruct. This spread indicates that the base autoencoders are learning diverse representations of the data, where CAE and LSTM-AE outperform vanilla AE, VAE, LSTM-VAE because they are adept at capturing spatial or sequential patterns in the data. This targeted strength is crucial for many tasks, allowing them to reconstruct data with higher fidelity compared to general-purpose autoencoders like vanilla or variational ones. Even though VAE's handle uncertainty, their focus might not directly benefit the reconstruction task in this case. By combining autoencoders that capture different aspects of the data, the ensemble has the potential to learn a more comprehensive representation, potentially leading to improved overall performance. This spread suggests diverse encoding capabilities within the ensemble.

*3.2.1 consensus-based results*

As per section 2.3.1, results from consensus of individual base-models using Eq. (5) are shown in Table 3.

Table 3: Results from Consensus method

| Base Model Names | Indexes of Common anomaly identified by all models | Number of anomalies ($N_a$) |
|---|---|---|
| AE, VAE, CAE, LSTM-AE, LSTM-VAE | 2020-12-09 10:02:00.702, 2020-12-09 10:02:01.734, 2020-12-09 10:02:02.765, 2020-12-09 10:02:03.796, 2020-12-09 10:02:04.827, 2020-12-09 10:02:05.859 | 6 |

By leveraging a consensus approach, the ensemble can combine the strengths of various autoencoder types (e.g., CAE for spatial features, LSTM for sequences) and improve anomaly detection accuracy. Though each autoencoder in the ensemble might flag different data points as anomalies. The consensus approach combines these outputs using *Eq. (5)*, to identify anomalies jointly and helps filter out false positives from individual models.

3.2.2 Hard /Majority Voting results

As per section 2.3.2, results from majority of the individual base-models using Eq. (6) are shown in Table 4.

Table 4: Results from Majority Voting

| identified Anomaly index | M1= AE | M2 = VAE | M3= CAE | M4= LSTM-AE | M5= LSTM-VAE | Final Prediction[a] |
|---|---|---|---|---|---|---|
| 2020-12-09 06:14:11.462 | 1 | 0 | 0 | 0 | 1 | 0 |
| 2020-12-09 06:14:12.493 | 0 | 1 | 0 | 0 | 1 | 0 |
| 2020-12-09 06:14:13.552 | 1 | 0 | 0 | 0 | 0 | 0 |
| 2020-12-09 10:01:59.671 | 1 | 0 | 1 | 1 | 0 | 1 |
| 2020-12-09 10:02:06.890 | 0 | 1 | 1 | 1 | 0 | 1 |
| 2020-12-09 10:02:07.921 | 1 | 0 | 1 | 1 | 0 | 1 |
| 2020-12-09 10:02:08.953 | 0 | 1 | 1 | 1 | 1 | 1 |
| 2020-12-09 10:02:10.499 | 0 | 1 | 0 | 0 | 1 | 0 |
| 2020-12-09 10:02:11.531 | 1 | 0 | 0 | 0 | 1 | 0 |
| 2020-12-09 11:03:04.358 | 0 | 0 | 0 | 0 | 1 | 0 |
| 2020-12-09 11:03:06.937 | 1 | 0 | 0 | 0 | 1 | 0 |
| 2020-12-09 11:08:47.779 | 1 | 0 | 0 | 0 | 0 | 0 |
| 2020-12-09 13:30:09.119 | 0 | 1 | 0 | 1 | 0 | 0 |
| 2020-12-09 13:30:10.151 | 1 | 1 | 0 | 0 | 0 | 0 |
| 2020-12-09 13:30:11.182 | 0 | 1 | 0 | 0 | 0 | 0 |
| 2020-12-09 13:30:12.729 | 0 | 1 | 0 | 1 | 0 | 0 |

[a] as per Eq. (6), Here 1 =anomaly, 0=normal. Thus, total number of anomalies given by Eq. (9) is $N_{b.1}$ =4.

Thus, voting-based ensemble of autoencoders shows promise, by combining reconstructions from multiple autoencoders, the approach likely captures a more robust representation of data, potentially leading to improved performance. It is important to note that the effectiveness of majority voting can depend on the size and diversity of the ensembles and that with a small number of similar autoencoders, the benefit might be minimal.

### 3.2.3 Weighted Average Voting results

As per section 3.2.2, first the weights $W_j$ for each model mae metric performances are derived using Eq. (7).

Table 5: Weights for weighted Average Voting

| Base-Model | Model Designation | Weights $W_j$ | Weight Designation | MAE score | $R_j$=performance Ranks | $RW_j$- Weights calculated[a] |
|---|---|---|---|---|---|---|
| AE | M1 | 0.57 | W1 | 0.43 | 3 | 0.2 |
| VAE | M2 | 0.547 | W2 | 0.453 | 2 | 0.133 |
| CAE | M3 | 0.993 | W3 | 0.007 | 5 | 0.333 |
| LSTM-AE | M4 | 0.884 | W4 | 0.116 | 4 | 0.266 |
| LSTM-VAE | M5 | 0.516 | W5 | 0.484 | 1 | 0.066 |

Now, from Table 5 the sum of the weights $W_j$ of the five models =3.51, applied onto the individual model predictions on anomaly indexes from Table 4 and using Eq. (8), final prediction is presented in Table 6.

Table 6: Results from Weighted Average Voting

| identified Anomaly index | M1= AE | M2 = VAE | M3= CAE | M4= LSTM-AE | M5= LSTM-VAE | Final Prediction[a] |
|---|---|---|---|---|---|---|
| 2020-12-09 06:14:11.462 | 0.57 | 0 | 0 | 0 | 0.516 | 0 |
| 2020-12-09 06:14:12.493 | 0 | 0.547 | 0 | 0 | 0.516 | 0 |
| 2020-12-09 06:14:13.552 | 0.57 | 0 | 0 | 0 | 0 | 0 |
| 2020-12-09 10:01:59.671 | 0.57 | 0 | 0.993 | 0.884 | 0 | 1 |
| 2020-12-09 10:02:06.890 | 0 | 0.547 | 0.993 | 0.884 | 0 | 1 |
| 2020-12-09 10:02:07.921 | 0.57 | 0 | 0.993 | 0.884 | 0 | 1 |
| 2020-12-09 10:02:08.953 | 0 | 0.547 | 0.993 | 0.884 | 0.516 | 1 |

| identified Anomaly index | *M1*= AE | *M2* = VAE | *M3*= CAE | *M4*= LSTM-AE | *M5*= LSTM-VAE | Final Prediction[a] |
|---|---|---|---|---|---|---|
| 2020-12-09 10:02:10.499 | 0 | 0.547 | 0 | 0 | 0.516 | 0 |
| 2020-12-09 10:02:11.531 | 0.57 | 0 | 0 | 0 | 0.516 | 0 |
| 2020-12-09 11:03:04.358 | 0 | 0 | 0 | 0 | 0.516 | 0 |
| 2020-12-09 11:03:06.937 | 0.57 | 0 | 0 | 0 | 0.516 | 0 |
| 2020-12-09 11:08:47.779 | 0.57 | 0 | 0 | 0 | 0 | 0 |
| 2020-12-09 13:30:09.119 | 0 | 0.547 | 0 | 0.884 | 0.516 | 1 |
| 2020-12-09 13:30:10.151 | 0.57 | 0.547 | 0 | 0 | 0 | 0 |
| 2020-12-09 13:30:11.182 | 0 | 0.547 | 0 | 0 | 0 | 0 |
| 2020-12-09 13:30:12.729 | 0 | 0.547 | 0 | 0.884 | 0 | 0 |

[a] as per Eq. (8), Here 1 =anomaly, 0=normal. As per Eq. (9), total anomalies $N_{b.2a}$ =5, here.

Here, the CAE has a higher weight based on its MAE score, while the LSTM-VAE has the lower weight. Each autoencoder's reconstruction contributes to the final result, but with a weight based on its performance of reconstruction error, as defined by Eq.(8). This allows models with better reconstructions to have a stronger influence, potentially improving the overall accuracy and robustness of the ensemble.

### 3.2.4 Rank-based Voting results

As per section 3.2.2, first the ranks $R_j$ for each model performances are assigned wherein worst performing model gets a rank of 1 and best performing model gets highest rank. Then using Eq. (10), the rank weights $RW_j$ are calculated, as shown in Table 5. The final prediction shown in Table 7 is derived using Eq. (11).

Table 7: Results from Ranking based weighted voting.

| identified Anomaly index | *M1*= AE | *M2* = VAE | *M3*= CAE | *M4*= LSTM-AE | *M5*= LSTM-VAE | Final Prediction[a] |
|---|---|---|---|---|---|---|
| 2020-12-09 06:14:11.462 | 0.2 | 0 | 0 | 0 | 0.066 | 0 |
| 2020-12-09 06:14:12.493 | 0 | 0.133 | 0 | 0 | 0.066 | 0 |
| 2020-12-09 06:14:13.552 | 0.2 | 0 | 0 | 0 | 0 | 0 |
| 2020-12-09 10:01:59.671 | 0.2 | 0 | 0.333 | 0.266 | 0 | 1 |
| 2020-12-09 10:02:06.890 | 0 | 0.133 | 0.333 | 0.266 | 0 | 1 |
| 2020-12-09 10:02:07.921 | 0.2 | 0 | 0.333 | 0.266 | 0 | 1 |
| 2020-12-09 10:02:08.953 | 0 | 0.133 | 0.333 | 0.266 | 0.066 | 1 |
| 2020-12-09 10:02:10.499 | 0 | 0.133 | 0 | 0 | 0.066 | 0 |
| 2020-12-09 10:02:11.531 | 0.2 | 0 | 0 | 0 | 0.066 | 0 |
| 2020-12-09 11:03:04.358 | 0 | 0 | 0 | 0 | 0.066 | 0 |
| 2020-12-09 11:03:06.937 | 0.2 | 0 | 0 | 0 | 0.066 | 0 |
| 2020-12-09 11:08:47.779 | 0.2 | 0 | 0 | 0 | 0 | 0 |
| 2020-12-09 13:30:09.119 | 0 | 0.133 | 0 | 0.266 | 0.066 | 0 |
| 2020-12-09 13:30:10.151 | 0.2 | 0.133 | 0 | 0 | 0 | 0 |
| 2020-12-09 13:30:11.182 | 0 | 0.133 | 0 | 0 | 0 | 0 |
| 2020-12-09 13:30:12.729 | 0 | 0.133 | 0 | 0.266 | 0 | 0 |

[a] as per Eq. (11) Here 1 =anomaly, 0=normal. As per Eq. (9), total anomalies $N_{b.2b}$ =4, here.

Thus, from Eq. (12), the overall total number of anomalies on the validation sample is calculated as follows:

$N = N_a + N_b$, where $N_a$ =6 and $N_b$ =majority of ($N_{b.1}$, $N_{b.2a}$, and $N_{b.2b}$) =4, N = 6 + 4 = 10

It is observed that even though the weighted average voting method has detected more anomalies as compared to the other 2 methods viz. majority and rank-based voting, based on discussion and confirmation with SME, this additional anomaly is invalidated. These 10 anomalies in this test dataset validated by domain experts are linked to the FMEA to cross link the key features prone to risk of the overall cooling system. Thus, in this practical application of Industrial cooling, to enhance reliability, reduce risks and minimize downtime the combination of different autoencoders through ensemble methods led to more accurate reconstructions of the original data than with a single best leaner (like CAE here). This could be beneficial in anomaly detection tasks, where identifying even slight deviations from normal patterns is crucial. The reliability and coverage is ensured by applying different methods to aggregate all the anomalies, which otherwise in ensemble anomaly detection methods as in [5], [6-8], would end up only with $N_a$, or one of $N_b$. The contribution of this experiment shows that the ensemble approaches can improve the overall robustness of the system, if one autoencoder struggles with a particular type of data, the others might compensate for its weaknesses, leading to more consistent performance, as seen in the result of the dual fusion approach and thus is the key motivation to adapt this strategy. This is despite the challenges encountered in the training computation time and managing the different architectures of the base-learners.

### 3.3 Data availability and reproducibility:

Due to privacy restrictions, the specific data that belongs to a real customer project used in this study cannot be publicly shared. While the code used in this research cannot be directly shared, anonymized code snippets demonstrating the core functionalities of the autoencoder ensembles are available upon request. Additionally, the implementation heavily relies on the deep learning frameworks like TensorFlow and Keras.

## 4 CONCLUSION

This work presents a novel approach to expand the horizon of anomaly detection for predictive maintenance in industrial cooling systems. By shifting from reactive schedules to anomaly-driven insights, we aim to significantly reduce downtime, enhance system reliability, and minimize costs. The unique contributions include utilizing diverse models to capture various failure patterns, a dual ensemble fusion of voting techniques for improved accuracy, and a focus on the specific intricacies of industrial cooling systems. As the industrial systems are dynamic and may experience changes over time due to various factors, such as operational adjustments or external influences, the dual fusion approach can adapt to such changes by continuously integrating information from multiple models and adjusting the fusion process accordingly. This adaptability ensures that the anomaly detection system remains effective even in evolving operating conditions. Furthermore, this approach directly complements traditional FMEA/FMECA processes by providing real-time anomaly detection beyond pre-identified failure modes. This proactive identification of emerging issues allows for targeted and timely interventions, preventing potential breakdowns and their associated costs. This approach can be easily generalized for other assets in the domain for the process of monitoring with enhanced horizon representation learning by different base learning models. While acknowledging limitations like data scope and model complexity, we envision expanding anomaly detection capabilities and seamlessly integrating this approach with existing maintenance practices. We believe this work has far-reaching implications, influencing numerous other assets within the power systems for extensive coverage in anomaly detection for predictive maintenance.

## ACKNOWLEDGMENTS

We acknowledge the support by the Swedish Foundation for Strategic Research (Project: ID20-0019).

**1. Please help us understand your paper better by completing below form, and it will not be published**

| First Author | Position : Ms. Sarala M Naidu |
|---|---|
| | Research Field: Self Supervised Learning methods for Predictive maintenance in Power systems |
| | Homepage URL: https://www.es.mdu.se/staff/4635-Sarala_Mohan |
| | |
| Second Author | Position: Prof. Ning Xiong |
| | Research Field: Learning and Optimisation |
| | Homepage URL: https://www.es.mdu.se/staff/82-Ning_Xiong |
| | |
| **Add more rows if necessary!** | |